%% file: techReport.tex
\documentclass[numbers,sort&compress]{article}
\usepackage[preprint]{neurips_2023}

\usepackage[utf8]{inputenc} 
\usepackage[T1]{fontenc}    
\usepackage{hyperref}       
\usepackage{url}            
\usepackage{booktabs}       
\usepackage{natbib}         
\usepackage{multicol}
\usepackage{amsfonts}       
\usepackage{nicefrac}       
\usepackage{microtype}      
\usepackage{xcolor}   
\usepackage{wrapfig}
\usepackage{multirow}
\usepackage{authblk}        
\usepackage{hyperref}       
\usepackage{url}            
\usepackage{booktabs}       
\usepackage{amsfonts}       
\usepackage{amssymb}        
\usepackage{nicefrac}       
\usepackage{microtype}      
\usepackage{lipsum}
\usepackage{relsize}
\usepackage{makecell}
\usepackage{enumitem}
\usepackage{colortbl}

\usepackage{fancyhdr}       
\usepackage{graphicx}       
\usepackage{caption} 
\usepackage{float}
\graphicspath{{media/}}

\newcommand{\gmidrule}{\arrayrulecolor{gray!50}\midrule\arrayrulecolor{black}}

\title{Yi: Open Foundation Models by 01.AI }

\author{
  \textbf{01.AI} \\
  \; \\ 
  \textbf{Code:}\; \url{https://github.com/01-ai/Yi}\\
  \textbf{Model:}\; \url{https://huggingface.co/01-ai} \\
}
\begin{document}
\maketitle

\begin{abstract}

%
We introduce the Yi model family, a series of language and multimodal models that demonstrate strong multi-dimensional capabilities. 
The Yi model family is based on 6B and 34B pretrained language models, then we extend them to chat models, 200K long context models, depth-upscaled models, and vision-language models. 
Our base models achieve strong performance on a wide range of benchmarks like MMLU, 
and our finetuned chat models deliver strong human preference rate on major evaluation platforms like AlpacaEval and Chatbot Arena.
Building upon our scalable super-computing infrastructure and the classical transformer architecture, 
we attribute the performance of Yi models primarily to its data quality resulting from our data-engineering efforts. 
For pretraining, we construct 3.1 trillion tokens of English and Chinese corpora using a cascaded data deduplication and quality filtering  pipeline. 
For finetuning, we polish a small scale (less than 10K) instruction dataset over multiple iterations such that every single instance has been verified directly by our machine learning engineers.
For vision-language, we combine the chat language model with a vision transformer encoder and train the model to align visual representations to the semantic space of the language model. 
We further extend the context length to 200K through lightweight continual pretraining and demonstrate strong needle-in-a-haystack retrieval performance. 
We show that extending the depth of the pretrained checkpoint through continual pretraining further improves performance. 
We believe that given our current results, continuing to scale up model parameters using thoroughly optimized data will lead to even stronger frontier models.

\end{abstract}


\clearpage
\tableofcontents
\clearpage

\section{Introduction}
\input{010_intro}




\section{Pretraining}
\label{sec:pretrain}

\input{020_pretraining}

\section{Finetuning}
\label{sec:finetuning}
\input{030_finetuning}

\section{Infrastructure}
\label{sec:infra}

\input{040_infra}

\section{Safety}
\label{sec:safety}
\input{031_safety}

\section{Evaluations}
\label{sec:evaluations}
\input{050_eval}

\section{Capability Extension}
In this section, we discuss our post-training methods to extend the Yi base model to 200K long-context, equip it with visual understanding capability, and enhance the 6B model by depth upsacaling.

\subsection{Long Context Modeling}
\label{sec:long_context}
\input{021_long_context}

\subsection{Vision-Language}
\label{sec:vision_language}
\input{022_vision_language}

\subsection{Depth Upscaling}
\label{sec:depth_upscaling}
\input{023_depth_upscaling}



\section{Final Discussions}
\label{sec:conclusion}
\input{070_conclusion}

\newpage
\appendix

\section{Author List and Contributions}
\input{app_author_list}


\clearpage
\bibliography{references}

\end{document}

%% file: 010_intro.tex
Recent breakthroughs in large language models have revolutionized the whole field of artificial intelligence and potentially radiate across the entire human society. 
Our vision for large language models is to make them the next generation computational platform and empower the whole community with significantly amplified intelligence. 
As a step towards this mission, we present the Yi model series, 6B and 34B language models pretrained from scratch on 3.1T highly-engineered large amount of data, and finetuned on a small but meticulously polished alignment data. 
Due to the data quality resulting from our substantial engineering efforts, which we will detail in the upcoming sections, Yi achieves near GPT-3.5 benchmark scores and human preferences. 

In designing the Yi model series, we are mostly concerned on the following dimensions regarding
\textit{model scale, data scale, and data quality}:
(1). when choosing model scale, the desiderata is to have small enough model that is feasible for inference on consumer-grade hardware like the RTX 4090 where the bounding factor is its limited 24G memory, yet still large enough with complex reasoning and emergent abilities. This is why we found 34B gives a nice performance-cost balance;
(2). since 34B is smaller than the conventional 70B used by Chinchilla~\citep{hoffmann2022training} and LLaMA~\citep{llama2}, we increase the pretrain data scale to 3.1T tokens to compensate for the decreased compute flops. This makes the model-data scale combination fall into the post Chinchilla optimal regime~\citep{sardana2023beyond}, i.e., we overtrain the model on more tokens (3T) than the compute optimal (around 1T). The benefit is from the inference side, as we achieve stronger performance with reduced serving cost: after int4~\citep{wu2023understanding} quantization, one can serve the 34B chat model on 24G GPU memory with almost no performance drop;
(3). our data engineering principle is to promote quality over quantity for both pretraining and finetuning.
The pretraining data quality is guaranteed by a sophisticated data cleaning pipeline with cascaded filtering methods and intentionally increased deduplication strength;
(4). for finetuning data we heavily emphasize quality by handcrafting less than 10K instructions over multiple iterations based on user feedback. This approach significantly deviates from the quantity-scaling styled instruction tuning works like FLAN~\citep{chung2022scaling} and UltraChat~\citep{ding2023enhancing}, but aligns more with the handcrafting styled works like LIMA~\citep{zhou2023lima}.

Our pretraining data cleaning system features a sophisticated filtering pipeline based on language, heuristic textual features, perplexity, semantics, topic, and safety, as well as a cascaded deduplication process based on paragraph, MinHash, and exact matching. 
This thorough pipeline leads to a much higher removal ratio than existing pipelines like CCNet~\citep{wenzek2019ccnet}, RefinedWeb~\citep{penedo2023refinedweb} and RedPajama~\citep{together2023redpajama}, which we believe is key to the success of data engineering.
The underlying principle is although pretraining requires data scaling, one would like to make sure the data used are of high quality, rather than training the model on large raw data, i.e.,  
we prefer 3T tokens over sophasticated engineering over 10T tokens without extensive filtering. 
Regarding the model architecture, we use standard implementation of the Transformer architecture with Grouped-Query Attention (GQA)~\citep{gqa}, SwiGLU~\citep{swiglu} activation, and RoPE with an adjusted base frequency (RoPE ABF)~\citep{xiong2023effective}.
This design choice is the standard approach rooted from the Transformer original paper~\citep{Vaswani:2017aa}, later modified by GPT-3 and Chinchilla~\citep{hoffmann2022training}, then followed by LLaMA~\citep{llama2}, Baichuan~\citep{Yang:2023aa}, Qwen~\citep{Bai:2023aa} and many related works.

To approach GPT-3.5-matching human preferences, our finetuning dataset is curated from carefully selected multi-turn instruction-response pairs, annotated directly by our team of machine learning engineers then polished over multiple iterations of user feedback.
As mentioned above, the size of our finetuning dataset is less than 10K, but improved over and over again across the model development timeline. 
Benefiting from the dataset's manageable size, we employed an extensive grid search to identify the optimal data composition, promote diversity, and discover effective hyperparameters. 
After 8-bit and 4-bit quantization, the final chat model can be deployed on consumer-grade GPUs nearly without performance degradation compared to the bf16 format.

We further extend the Yi model capability from three dimensions: context scaling, vision-language adaptation, and depth-upscaling. 
To achive 200K context length, we continue pretrain the model on about 5B length-upsampled data, similar to the concurrent work in~\citet{fu2024data}. 
To adapt the model to vision-language tasks, we integrate a vision encoder and develop a multi-stage training method, following and improving the practice of~\citet{liu2023visual}. 
We also study the effectiveness of depth-upscailng~\citep{kim2023solar}, i.e., making the model deeper by continual pretraining, and confirming its effectiveness to further improve model performance.

Our infrastructure provides strong support for the full-stack development of the Yi model series, from pretraining to finetuning to serving. 
To support pretraining, we develop cross-cloud elastic task scheduling, automatic failure recovery, and topology-aware resource allocation which collectively enable us to run tasks according to the real-time available GPU nodes cross clusters with limited switching overhead. 
To support finetuning, we build a hierarchical scheduling framework supporting different distributed backends for different models (e.g., Megatron~\citep{megatron-lm} for the policy model and DeepSpeed~\citep{zero} for the reward model). 
For efficient inference, we use 4-bit model and 8-bit KV cache quantization, combining with PagedAttention~\citep{kwon2023efficient} and Dynamic Batching.

Extensive experiments demonstrate that Yi-34B can match GPT-3.5 in both performance and efficiency. 
On most standard benchmarks like MMLU~\citep{hendrycks2020measuring} (for the base model) and LMSys ELO Rating~\citep{zheng2023judging} (for the chat model), Yi-34B generally achieves scores on par with GPT-3.5. 
After model parameter and KV cache quantization, the inference cost is also controlled such that a wide range of the community can deploy the model on cost effective devices. 
We further report a detailed performance comparison between Yi and major LLMs on commonsense reasoning, college exams, math, coding, reading comprehension, and human preference win-rate on multiple evaluation benchmarks.

Since its release, the Yi model series has  benefited the community from the following perspectives: 
(1). it provides GPT-3.5-matching quality yet cost-effective models to researchers, and enables developers to build AI-native applications like language model based agents;
(2). it empowers end users with locally runnable chatbots, which consequently helps protecting user data privacy; 
(3). it sheds light on the direction on further data and model scaling to achieve even stronger frontier models. 
for both research and commercial use.

%% file: 020_pretraining.tex
\input{images/fig_data_pipeline}

Our approach to pretraining is to train a standard dense transformer architecture on a heavily engineered large pretraining corpora, where our underlying assumption is that when trained on extensive data of high-enough quality, a standard architecture can exhibit advanced capability. 
This is to say, we may not need much architectural modification, although we have indeed conducted extensive preliminary architectural experiments. 
In the following subsections, we first detail our data engineering pipeline, then briefly discuss the model architecture. 

\subsection{Data Processing}



The Yi data mixture is shown in Fig.~\ref{fig:data_mixture}.
To produce a high-quality bilingual pretraining data, 
we meticulously designed a cascaded data-processing pipeline, as illustrated in Fig~\ref{fig:data_cleaning}. 
This pipeline features a series of data-cleaning strategies targeting  quality and diversity. 
We start with web documents from Common Crawl, use the CCNet pipeline \citep{Wenzek:2019aa} for language identification and perplexity  scoring. 
Then we use a combination of filtering and deduplication process, as detailed below. 



\paragraph{Heuristic Rule Filters} 
This part of filter aims for removing text of low quality. 
We filter out text based on: 
(1). URL, domain, word blocklists and garbled text filters; 
(2). document length, the ratio of special symbols, and the ratio of short, consecutive, or incomplete lines;
(3). repeated words, n-grams, or paragraphs \citep{rae2021scaling};
The filtering thresholds are based on a statistical analysis of large document samples, as described in \citet{nguyen2023culturax}. 
Furthermore, we identify and anonymize Personal Identifiable Information (PII), such as email addresses and phone numbers.

\input{images/fig_data_mix}
\paragraph{Learned Filters} 

We use learned filters to address nuanced cases that exceed the capabilities of standard heuristic rules. Notably, the Chinese content extracted from Common Crawl present unique challenges, particularly with a higher ratio of inappropriate content like pornography and gambling. Traditional heuristic-rule-based filters struggle to effectively identify and eliminate all harmful content. To enhance our filtering process, we have integrated a suite of learned scorers for filtering, namely the perplexity scorer, quality scorer, safety scorer, and document coherence scorer:
(1). the \textit{Perplexity Scorer}, utilizing the KenLM library as per CCNet \citep{wenzek2019ccnet}, evaluates a vast array of web documents, discarding those with perplexity scores largely above average;
(2). the \textit{Quality Scorer} is a classifier trained to recognize and favor pages similar to Wikipedia in quality and assign scores accordingly. Documents that fail to meet the quality standard are subsequently removed;
(3). the \textit{Document Coherence Scorer} identifies low-quality web documents that consist of disparate sentences or paragraphs, thus being incoherence. Such documents are either segmented for further analysis or removed entirely. 
(4). the \textit{Safety Scorer} identifies and removes web documents containing toxic content, such as violence, pornography, and political propaganda.

\paragraph{Cluster-based Filters} 
We further use unsupervised semantic clustering to group web documents. This clustering process enables efficient identification and analysis of documents sharing similar semantic features. The clustered data are subsequently annotated with quality labels, providing essential references for the optimization of Yi's data mixture strategy. Documents identified as low-quality through automatic and manual verification are excluded from the dataset.

\paragraph{Deduplication} 
After filtering, we implement a comprehensive deduplication pipeline following the procedure in Penedo et al. (2023) \citep{penedo2023refinedweb}. This pipeline integrates document-level MinHash deduplication and sub-document exact-match deduplication, effectively identifying and removing duplicate content within and across documents. We further categorize web documents into specific themes using a topic model predicting labels like as news, ads, and knowledge-based content. In the final pretraining dataset, we down-sample less helpful content, mostly advertisements, to ensure information density. The final composition of Yi's pretraining data  is shown in Fig.~\ref{fig:data_mixture}.

\subsection{Tokenization}
We use byte-pair encoding (BPE)~\citep{shibata1999byte} implemented in the SentencePiece framework~\citep{kudo2018sentencepiece}, to tokenize the pretraining data.
The vocabulary size of Yi is set to 64,000 to balance computational efficiency and word comprehension.
Specifically, we split numbers into individual digits to facilitate a better understanding of numeric data.
We allow rare characters to fall back to the unicode-byte encoding to ensure fault tolerance.
We employ the identity tokenizer to avoid transferring all punctuations to the half-width format.
LLMs prioritizing English usually utilize dummy prefix (whitespace at the beginning of text) in their tokenizers to generalize the same words at different positions of sentences.
We do not use this approach because the assumption does not always hold even in the English context, especially for sentences that begin with quotation marks, also it does not show positive effect in Chinese context.

\subsection{Model Architecture}
\begin{table}[t!]
    \centering
    \begin{tabular}{@{}cccccccc@{}}
        \toprule
         Models&  Hidden Size&  Q-heads&  KV-heads&  Layers&  Pretrain Seq. Len& Max LR\\
         \midrule
         6B&  4096&  32&  4&  32&  4096& $3\times10^{-4}$\\
         34B&  7168&  56&  8&  60&  4096& $1.5\times10^{-4}$\\
         \bottomrule
    \end{tabular}
    \space
    \caption{Model configs of Yi-6B and Yi-34B. LR stands for learning rate.}
    \label{tab:model_config}
\end{table}
Yi uses a modified version of the classical decoder-only Transformer architecture~\citep{Vaswani:2017aa} where the code is based on LLaMA's~\citep{llama2} implementation. 
The main parameter setting is summarized in Table~\ref{tab:model_config}. 
The modifications from LLaMA to Yi are further summarized below:

\paragraph{Attention Mechanism} LLaMA 2 uses Grouped-Query Attention(GQA)~\citep{gqa} only on its largest 70B model, and its 7B and 13B uses full attention. We incorporate GQA in both Yi-6B and Yi-34B. 
GQA splits query-heads into G groups, sharing a single key and value head within each group of query~\citep{gqa}.
This approach offers substantial reductions of training and inference costs, compared to the original Multi-Head Attention (MHA)~\citep{shazeer2019fast,de2022fido,pope2023efficiently}.
We do not observe performance degradation after applying GQA to our 6B smaller model. 


\paragraph{Activation Function} 
We use SwiGLU~\citep{swiglu} as Yi's post-attention layer, reducing its activation size from $4h$ to $8/3h$ ($h$ denotes hidden size) to be consistent with the normal post-attention layer. 
This adjustment also compensates for the reduction in parameter resulted from GQA, making the overall parameter count comparible of existing 7B and 34B models. 


\paragraph{Positional Embedding and Long Context}

We use Rotary Position Embedding 
 (RoPE)~\citep{su2021roformer} following the standard implementation.
We adjust the base frequency (RoPE ABF), introduced in~\citet{xiong2023effective}, to support long context windows up to 200K where the base model itself is trained on 4K context length. 
To adapt the base model to longer context, we continue pretrain the model on 10B tokens from our pretraining data mixture with slightly upsampled long sequences, mostly from book. 
We observe that only 1-2B tokens is enough for the model to converge to low loss on 4K-200K length, and a lightweight finetuning further induces near-perfect long-context retrieval performance. 
Based on this observation, we tend to view that the capability of modeling longer dependency than the pretrained length (4K) is a intrinsic capability (rather than an being injected by post-train).
This is to say, the base model already has the capability to model longer than 4K dependency even the model is trained shorter, and the post-train / finetuning procedure simply release this capability.



%% file: images/fig_data_pipeline.tex
\begin{figure}
  \centering
  \includegraphics[scale=0.18]{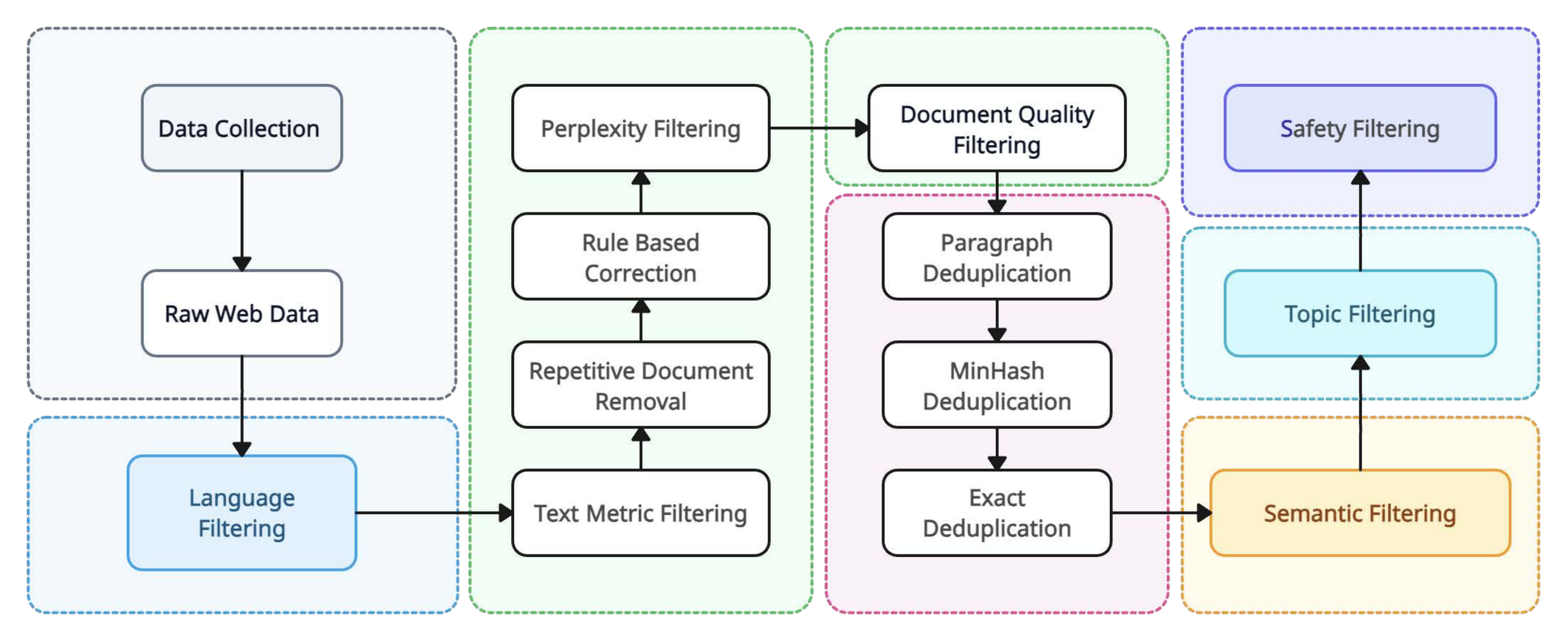}
  \caption{Yi's pretraining data cleaning pipeline. 
  }
  \label{fig:data_cleaning}
\end{figure}

%% file: images/fig_data_mix.tex
\begin{wrapfigure}{r}{0.5\textwidth}
	\begin{center}
		\vspace{-0.3in}
		\includegraphics[width=0.5\textwidth]{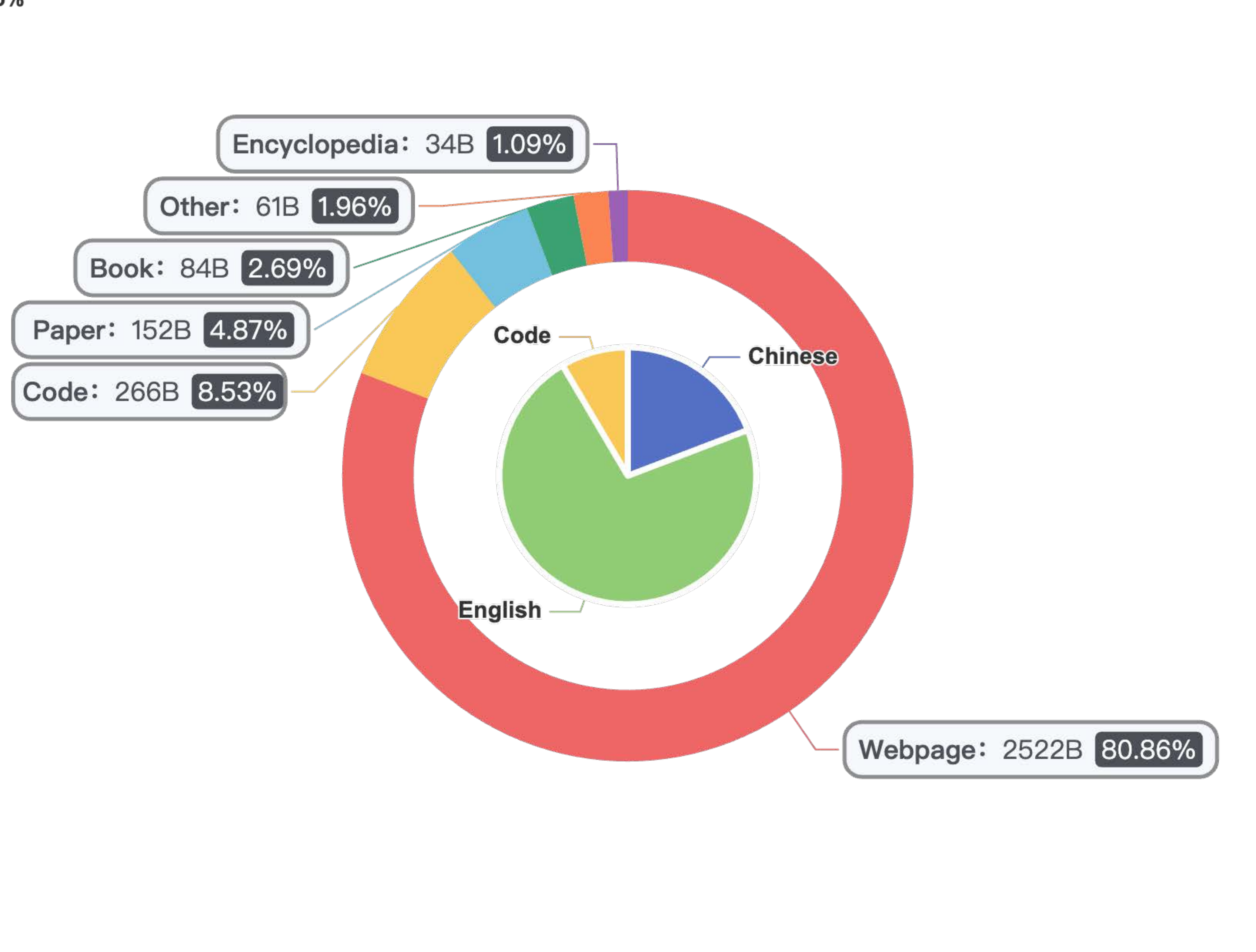}
	\end{center}
	\vspace{-0.15in}
	\caption{Yi's pre-training data mixture. Overall our data consist of 3.1T high-quality tokens in Both English and Chinese, and come from various sources. Our major differences from existing known mixtures like LLaMA~\citep{touvron2023llama} and Falcon~\citep{penedo2023refinedweb} are that we are bilingual, and of higher quality due to our more rigorous cleaning pipeline.} %
	\label{fig:data_mixture}
    \vspace{-0.2in}
\end{wrapfigure}

%% file: 030_finetuning.tex

Our finetuning method significantly emphasizes data quality over quantity.
Our approach does \textit{not} follow existing data-intensive approaches like FLAN~\citep{chung2022scaling} and UltraChat~\citep{ding2023enhancing}, which scales the SFT data to millions of entries but each of the entries may not been examined carefully because the scale is too large.
In contrast, our method aligns with the LIMA~\citep{zhou2023lima} and DEITA~\citep{liu2023makes} approach, which focus on data selection rather than scaling. 
With the scale being less than 10K, we are able to examine and optimize \textit{every single data point}. 
Below we discuss our data construction and training details.

\subsection{Data Preprocessing}

\paragraph{Quality is All You Need}
Our finetuning dataset consists of less than 10K multi-turn instruction-response dialog pairs, with each and every one of the entry constructed and polished over multiple iterations and from user feedback.
We take this approach because  in our preliminary experiments, we observe that compared to the open-source data of several hundred thousand entries, the results from a smaller, manually annotated dataset are superior.
These observations align with those reported in~\citet{llama2, zhou2023lima, team2023gemini}. 

We use the following techniques to improve prompt distribution selection, response formatting, and chain-of-thought formatting:
(1). for prompt distribution selection,
drawing inspiration from WizardLM\citep{xu2023wizardlm}, we develope compound instructions and progressively evolved them to increase their complexity. 
This approach has significantly reduced the size of SFT data 
in our experiments;
(2). for response formatting, we generally use a default style extended from LIMA\cite{zhou2023lima}. Overall, the responses are structured in an introduction-body-conclusion format where the body is usually a list of bullet point;
(3). for CoT data formatting, we have use a ``Step-Back'' pattern, inspired by \citet{zheng2023step}, by performing abstraction to formulate higher-level solutions before delving into reasoning about the original, more concrete questions.

We spend extra efforts on reducing hallucination and repetition:
(1). to reduce hallucinations, 
we examine and ensure that the knowledge in the responses is not contained within the model, and eliminate responses that might lead to memorization;
(2). to reduce repetition, we rewrite the repetitive turns of the responses that usually exist but may be overlooked in the finetuning data. 


\paragraph{Diversity and Mixture}
To ensure the coverage of different capabilities, we have included a wide spectrum of open-source prompt,
encompassing areas such as question answering, creative writing, dialogue, reasoning, mathematics, coding, safety, bilingual capabilities, and others. 

To obtain a fine-grained control of different directions of capabilities, inspired by InsTag\citep{lu2023instag}, we develop a instruction tagging system. By designing a diversity-focused sampling algorithm, we carefully balanced the distribution of instructions across various tags. This approach ensures a diverse finetuning dataset, aiming to achieve enhanced cross-task robustness.

To achieve the optimal data ratio for balancing different directions of the capability, we use an approximate grid search to determine our data mixture. Motivated by \citet{dong2023abilities}, this process involved experimenting with \{1, 1/2, 1/4, 1/8, 1/16, 1/32, 1/64\} proportions for each ability. The search process was guided by validation results 
and our in-house human evaluation sets.

\paragraph{ChatML Format} Beyond the focus on data quality and diversity, our observations revealed that the format of the data substantially influences the model's ultimate performance. To this end, we implemented the ChatML-style format~\citep{chatml}. This structured approach empowers the model to differentiate among various information types, such as system configurations, user inputs, and assistant responses.

\subsection{Training Method}

We use next-word prediction loss for finetuning, and only compute loss on the responses, but not system and user instructions. 
We use AdamW optimizer
with $\beta_1$ set to 0.9, $\beta_2$ set to 0.999, and $\epsilon$ set to $10^{-8}$. 
We use a sequence length of 4096, alongside a batch size of 64.
We set training step to 300 with a constant $1\times 10^{-5}$ learning rate, a weight decay of 0.1, gradient clipping with a maximum threshold of 1.0, and NEFTune~\cite{jain2023neftune} with a noise scale of 45 for Yi-34B-Chat and 5 for Yi-6B-Chat.

%% file: 040_infra.tex
We build the infrastructure supporting the full-stack data processing, pretraining, finetuning, and serving. Our infrastructure feasures:
(1). automated managing and monitoring the computing resource; 
(2). improved the training speed from optimized parallel strategies, kernel efficiency, and long-context support;
(3). unified finetuning framework supporting heterogeneous distributed training backend, such as simultaneously using Megatron and DeepSpeed for multiple models in Direct Preference Optimization (DPO)~\citep{rafailov2023direct};
(4). reducing the deployment cost by various LLM serving accelerations such as quantization, continuous batching, and paged attention.
Below we explain these techniques one by one. 

\paragraph{Computing Resources Management} 
To efficient schedule large-scale language model development, particularly pretraining, which may take months on thousands of GPUs , 
we build a highly efficient multi-cloud task scheduling algorithm to manage pre-training, SFT, and RLHF tasks of different priorities. We also build a high-performance in-house training framework that allows us to automatically elastic scale the pre-train jobs to different node sizes based on the GPU availability. More importantly, all the training-related hyper-parameters will be scaled at the same time seamlessly. 

During the large language model training stage, a wide range of failures regularly occur, ranging from GPU crashes to communication fabric errors to loss spikes. We use the following strategies to address these reliability challenges:
(1) we apply automated inspection, prediction, and labeling of nodes for different kind of software/hardware error categories. Nodes marked as tainted will be temporarily removed from the resource pool until the errors got cleared. 
(2) we implement a task queuing system with pre-checks and the capability for fast, automatic recovery in the event of failures during training tasks. 
(3) we develop of a user-friendly multi-task submission and management console, enabling developers to seamlessly manage and track their training tasks and hyper-parameters.

\paragraph{Performance and Cost Efficiency} \textit{Memory} and \textit{communication} restrictions are the two major technical challenges of large scale model training requiring integrated solutions beyond adding more GPUs.
We use and improve upon the following techniques to tackle the memory and communication restrictions: (1) ZeRO-1~\cite{zero} to remove the memory consumption by partitioning optimizer states cross data-parallel processes; (2) tensor parallel combined with pipeline parallel~\cite{megatron-lm} within each compute node to avoid inter-node communication bottleneck, and the 3D parallel strategy is well designed and optimized to avoid using activation checkpointing and minimize the pipeline bubbles; (3) kernel fusion techniques like flash attention\cite{dao2022flashattention}\cite{dao2023flashattention2} and JIT kernels to reduce redundant global memory access and consumption; (4) topology-aware resource allocation (ranking strategy) to minimize the communication across different layers of switches, which is the limitation of a typical fat-tree-topology. 

\paragraph{Finetuning Framework}
Different from pretraining, finetuning LLMs may require the orchestration of multiple models, as is the practice of DPO~\citep{rafailov2023direct} and PPO~\citep{ouyang2022training}.
In such training jobs, a typical process is to use reference/reward model to predict a batch of data (which also requires nontrivial time), then let the target model use this data to calculate loss and update parameters. 
To this end, we build a multi-model scheduling framework to support multiple backends for different LLMs in a single job. For example, when finetuning a language model with DPO, the intermediate results from the reference model can be cached and reused, improving the training speed and resource cost to be close to the supervised finetuning counterparts.

\input{tables/tab_overall}

\paragraph{Fast and Efficient Inference}
We primarily use quantization, dynamic batching, and Paged Attention for improving decoding speed and memory usage.
We use quantization to decrease both the memory footprint and computation demand. 
By 4-bit model quantization~\citep{wu2023understanding} and 8-bit KV cache quantization~\citep{dettmers2022llm}, we are able to achieve significant GPU memory saving with near-zero performance degradation (e.g., less than 1$\%$ accuracy drop in MMLU/CMMLU benchmark). 
We use dynamic batching~\citep{gyeong280922} to minimize the response time and improve batching efficiency.
We use PagedAttention\cite{kwon2023efficient} to improve memory utilization and improve decoding. 



\paragraph{Long-context Window Support} 
We implement and improve computation-communication overlapping, sequence parallelism, and communication compression to support up to 200K context length continue pretraining and finetuning. 
Our method to scale the context length to 200K is \textit{solely} based on engineering, that is to say, we do not modify the model architecture like sparse, local, or sliding window attention -- the model remains using the full attention even the input is 200K.  


%% file: tables/tab_overall.tex
\begin{table*}[tb!]
\centering
\scalebox{0.9}{
\begin{tabular}{@{}l c c c c c c c c c c c@{}}
\toprule
\multicolumn{1}{l}{\multirow{2}{*}{\textbf{}}}  & \multicolumn{1}{c}{\multirow{2}{*}{\textbf{Size}}} & \multicolumn{1}{c}{\multirow{2}{*}{\textbf{MMLU}}}  & \multicolumn{1}{c}{\multirow{2}{*}{\textbf{BBH}}} & \multicolumn{1}{c}{\multirow{2}{*}{\textbf{C-Eval}}} & \multicolumn{1}{c}{\multirow{2}{*}{\textbf{CMMLU}}} & \multicolumn{1}{c}{\multirow{2}{*}{\textbf{Gaokao}}} & \multicolumn{1}{c}{\multirow{2}{*}{\textbf{CR}}} & \multicolumn{1}{c}{\multirow{2}{*}{\textbf{RC}}} & \multicolumn{1}{c}{\multirow{2}{*}{\textbf{Code}}} & \multicolumn{1}{c}{\multirow{2}{*}{\textbf{Math}}} \\  
\multicolumn{1}{c}{} & \multicolumn{1}{c}{} & \multicolumn{1}{c}{} & \multicolumn{1}{c}{}    & \multicolumn{1}{c}{}  & \multicolumn{1}{c}{} & \multicolumn{1}{c}{} &  \multicolumn{1}{c}{} &  \multicolumn{1}{c}{}  &  \multicolumn{1}{c}{} \\
\midrule
\textbf{GPT-4}
& - & \textbf{83.0} & \textbf{86.7} & 69.9 & 71.0 & 72.3 & \textbf{89.3} & - & \textbf{65.3} & \textbf{66.1} \\
\textbf{GPT-3.5}
& - & 69.1 & 70.1 & 52.5 & 55.5 & 51.1 & 83.1 & - & 54.8 & 35.6 \\
\gmidrule
\multirow{1}{*}{\textbf{Qwen}} 
& 14B & 66.7 & 53.4 & 72.1 & 71.0 & 62.5   & 74.2 & 72.5 & 40.6 & 43.1 \\
\gmidrule
\multirow{2}{*}{\textbf{Llama2}} 
& 34B & 62.6 & 44.1 & - & - & -  & 71.1 & 68.9 & 27.8 & 24.2 \\
& 70B & 69.7 & 64.9 & 50.1 & 53.3 & 23.3 & 72.7 & 72.3  & 38.4 & 35.2  \\
\gmidrule
\multirow{1}{*}{\textbf{Baichuan-2}} 
& 13B & 55.0 & 49.0 & 59.0 & 61.97 & 45.6 & 66.3 & 62.4  & 23.4 & 16.1 \\
\gmidrule
\textbf{InternLM} & 20B & 62.1 & 52.5 & 58.8 &  59.0  & 45.5 & 78.3 & - & 34.8 & 30.26 \\
\gmidrule
\textbf{Skywork} & 13B & 62.1 & 41.7 & 60.6 & 61.8 & 68.1  & 72.4 & 61.4 & 64.9 & 18.1   \\
\gmidrule
\textbf{Falcon} & 180B &  70.4  &  54.0  &  57.8  &  58.0  &  59.0 &  74.4  &  - &  -    &  -    \\
\gmidrule
\multirow{2}{*}{\textbf{Yi}} & 6B & 63.2 & 42.8 & 72.0 & 75.5 & 72.2 & 72.2 & 68.7 & 21.1 & 18.6  \\
& 34B & 76.3 & 54.3 &\textbf{81.4} & \textbf{83.7} & \textbf{82.8} & 80.7 & \textbf{76.5}  & 32.1 & 40.8   \\

\bottomrule

\end{tabular}
}
\caption{Overall performance on grouped academic benchmarks compared to open-source base models. \textbf{CR} stands for Commonsense Reasoning. \textbf{RC} stands for Reading Comprehension.
}
\label{tab:models}
\end{table*}

%% file: 031_safety.tex
To enhance the model's trustworthiness and safety, we develop a full-stack Responsible AI Safety Engine (RAISE). 
RAISE ensures safe pretraining, alignment, and deployment.
This section discusses our safety measures in the pretraining and alignment stages.

\paragraph{Safety in Pretraining}
Aligning with standard pretraining data safety practices~\citep{llama2, rae2021scaling, brown2020language}, we build a set of filters based on heuristic rules, keyword matching, and learned classifiers
to remove text containing personal identifiers and private data, and reduce sexual, violent, and extremist content. 

\paragraph{Safety in Alignment}
Informed by existing research in~\citep{glaese2022improving, ji2023beavertails}, we first build a comprehensive safety taxonomy. 
This taxonomy covers a broad spectrum of potential concerns, including environmental disharmony, superstitious, religious sensitivities, discriminatory practices, substance abuse, violent behavior, illegal activities, hate speech, ethical violations, privacy breaches, self-harm, sexually explicit content, mental health issues, and cybersecurity threats. We curated datasets reflecting these categories for a robust alignment, and mix them with our dialog SFT data.
We also include a targeted set of prompts simulating attack scenarios in the alignment phase, which effectively improved the model's resilience against malicious use.

%% file: 050_eval.tex
Our evaluation demonstrates that the Yi model family achieves inspiring performance on a wide range of tasks and delivers close to GPT-3.5 user preference rate. We first report the base model performance on standard benchmarks, then we discuss the chat model performance and its user preference rate.

\subsection{Base Model Performance}
\input{051_eval_base}

\subsection{Chat Model Performance}
\input{051_eval_chat}



%% file: 051_eval_base.tex
\subsubsection{Main Results}
\label{sec:main results}

Here we present the results for our base models and several other well-known base models across standard academic benchmarks. While benchmarking open-source models, we observed a disparity between the results generated by our pipeline and those reported in public sources. Upon conducting a more in-depth investigation of this difference, mostly because different models use different prompts, post-processing strategies, and sampling techniques.
These differences may potentially induce significant variations in the outcomes. Our prompt and post-processing strategy remains consistent with the default settings of the original benchmarks\citep{PIQA,siqa,hellaswag,winogrande,arc,obqa,csqa,squad,quac,boolq,cobbe2021training,hendrycks2021measuring,chen2021evaluating,austin2021program,hendrycks2020measuring,li2023cmmlu,zhang2023evaluating,srivastava2023imitation,suzgun2022challenging}.
We use greedy decoding without any post-processing for the generated content. For scores that were not reported publicly (or scores reported with different settings), we try to get results with our pipeline. For scores that can be found publicly, we directly report the existing numbers. We use the following benchmarks, largely following the practice of LLaMA 2~\citep{llama2}:

\begin{description}[leftmargin=0.2in]
    \item[Commonsense Reasoning:] We included PIQA\cite{PIQA}, SIQA\cite{siqa}, HellaSwag\cite{hellaswag}, WinoGrande \cite{winogrande}, ARC\cite{arc}, OpenBookQA(OBQA)\cite{obqa}, and CommonsenseQA(CSQA)\cite{csqa} to assess common sense reasoning. CSQA was exclusively tested using a 7-shot setup, while all other tests were conducted with a 0-shot configuration.
    \item[Reading Comprehension:] For reading comprehension, we report the 0-shot average on SQuAD\cite{squad}, QuAC\cite{quac}, and BoolQ\cite{boolq}.
    \item[Math:] We report the average of the GSM8K\cite{cobbe2021training} (8 shot), and MATH\cite{hendrycks2021measuring} (4 shot) benchmarks with pass@1 accuracy without any specific prompting strategy (e.g. Chain-of-Thought prompting) and other ensemble technique (e.g., majority voting).
    \item[Code:] We report the average pass@1 scores of our models on HumanEval\cite{chen2021evaluating}  (Chen et al., 2021) and MBPP\cite{austin2021program} (Austin et al., 2021). 
    \item[Popular Aggregated Benchmark:] We report the overall results for MMLU\cite{hendrycks2020measuring}(5-shot), CMMLU\cite{li2023cmmlu} (5-shot), Gaokao-Bench\cite{zhang2023evaluating} (5-shot), and BigBench\cite{srivastava2023imitation} Hard (BBH\cite{suzgun2022challenging}) (3-shot).
\end{description}

\input{tables/tab_math}

By training on a significantly larger number of tokens (3.1T) compared to prior work (usually $\le$ 2T), we have observed a substantial performance gain across  benchmarks, as shown in Table \ref{tab:models}. However, it is important to note that there are still discernible disparities between our model and existing open-source and close-source models, particularly in tasks related to mathematics and coding. As performance in these domains can be significantly improved by continual pretraining and instruction fine-tuning, we have refrained from incorporating extensive mathematical and coding content in the pretraining corpus when making the initial design choices. We do plan to release models with enhanced math and coding capabilities in the future.

\input{images/fig_icl}

\subsubsection{Discussions}

\begin{description}[leftmargin=0.2in]
\item[Gain from Model Scale.] We observe that Yi-34B has substantial performance improvement compared to Yi-6B, though they utilized the same pretrain corpora.
Larger model size leads to higher performance gain on Code and Math benchmarks, referring to Tab.~\ref{tab:code_math}, compared to benchmarks focusing on Commonsense Reasoning, Reading Comprehension, or Knowledge.
\item[Data Quality.]
Smaller models of higher quality pretrain data, like Yi-34B or Qwen-14B, usually demonstrate better performance than models of larger size but (presumably) lower quality data, such as Falcon-180B (though the focus of Falcon-180B might be more on the scaling side, which is definitely of important value on its own). 
\item[Gap between GPT-4 and Open-source LLMs.]
Based on Tab.~\ref{tab:models}, we note that open-source LLMs still lag behind the performance of GPT-4 and GPT-3.5 on various benchmarks.
Yet representative bilingual LLMs, e.g. Qwen-14B and Yi-34B, can match or even surpass the performance of GPT-4 on Chinese knowledge related benchmarks, including C-Eval~\cite{huang2023c}, CMMLU~\cite{li2023cmmlu}, and Gaokao~\cite{zhang2023evaluating}.
However, there is still a huge gap between GPT-4 and open-source models on reasoning-related benchmarks like BBH~\cite{srivastava2023imitation}, code (HumanEval), and math (MATH).
\end{description}

\subsubsection{In-Context Learning Study}
We further investigate the in-context learning capability, i.e., the capability of inferring the underlying function given the few-show input-output demonstrations.
We consider the task of inferring the linear coefficient of a weighted sum.
Specifically, define $y = w_1x_1 + w2x_2 + ... + w_nx_n$, our few-shot demonstration is $x_1, x_2, ..., x_n, y$, and we ask the model to (implicitly) infer $w_1, w_2, ..., w_n$ by predicting the $y$ given a new set of input $x$. 
We use (a). the absolute difference between model prediction $y$ and the ground truth $y^*$, i.e., $|y - y^*|$ as a continuous measure, and use (b). the exact match $y == y^*$ as a discontinuous measure.
We further note that most of the models perform reasonably well on addition and subtraction, so the ability to do arithmetic, as a confounding factor, can be ruled out. 

The results are shown in Figure~\ref{fig:icl}.
When setting the linear coefficients of be [1, -1], we see that Yi 34B and LLaMA-2 70B performs the best in-terms of answer exact match. 
If we increase the number of the linear coefficients to be [1, 1, 1, 1, 1], we observe the emergent behavior that only large models (LLaMA-2 70B and Mixtral) can achieve good scores on exact match, although the differences to target is more continuous. 
These observations give side evidence for Yi-34B's performance on in-context learning and indicates that further scaling may allow the model to infer more complicated functions by in-context learning.

%% file: tables/tab_math.tex
\begin{table}
  \centering
  \begin{tabular}{@{}lccccc@{}}
    \toprule
    \bf Model & \bf Size & \bf GSM8k & \bf MATH & \bf Human-Eval pass@1 & \bf MBPP pass@1 \\
    \midrule
    \bf GPT-3.5 & - & 57.1 & 14.0 & 48.1 & 61.4 \\
    \bf GPT-4 & - & \textbf{92.0} & \textbf{40.2} & \textbf{67.0} & \textbf{63.6} \\
    \gmidrule
    \bf Falcon & 180B  & 54.4 & - & 0.61 & 47.0 \\
    \gmidrule
    \multirow{2}{*}{\bf Qwen}
    & 7B  & 51.7 & 11.6 & 29.9 & 34.0 \\
    & 14B  & 61.3 & 24.8 & 32.3 & 48.9 \\
    \gmidrule
    \multirow{2}{*}{\bf Baichuan 2}
    & 7B & 24.5 & 5.6 & 18.3 & 28.3 \\
    & 13B & 22.1 & 10.1 & 20.7 & 26.1 \\
    \gmidrule
    \multirow{3}{*}{\bf L\textsmaller{LaMA} 2} 
    & 7B  & 16.7 & 3.3 & 12.8 & 14.8\\ 
    & 34B  & 42.2 & 6.2 & 22.6 & 33.0\\ 
    & 70B  & 56.8 & 13.5 & 31.7 & 45.0\\
    \gmidrule
    \bf Mistral & 7B & 47.5 & 11.3 & 30.5 & 47.5 \\
    \gmidrule
    \bf InternLM & 20B & 62.9 & 10.9 & 28.1 & 41.4 \\
    \gmidrule
    \bf Skywork & 7B & 55.8 & 7.8 & 13.4 & 22.8 \\
    \gmidrule
    \multirow{2}{*}{\bf Yi}  
    & 6B & 32.5 & 4.6 & 15.9 & 26.3 \\
    & 34B & 67.2 & 14.4 & 23.2 & 41.0 \\
    \bottomrule
  \end{tabular}
  \caption{Comparison of models on GSM8k, MATH, Human-Eval, and MBPP.}
  \label{tab:code_math}
\end{table}

%% file: images/fig_icl.tex
\begin{figure}
	\begin{center}
		\includegraphics[width=\textwidth]{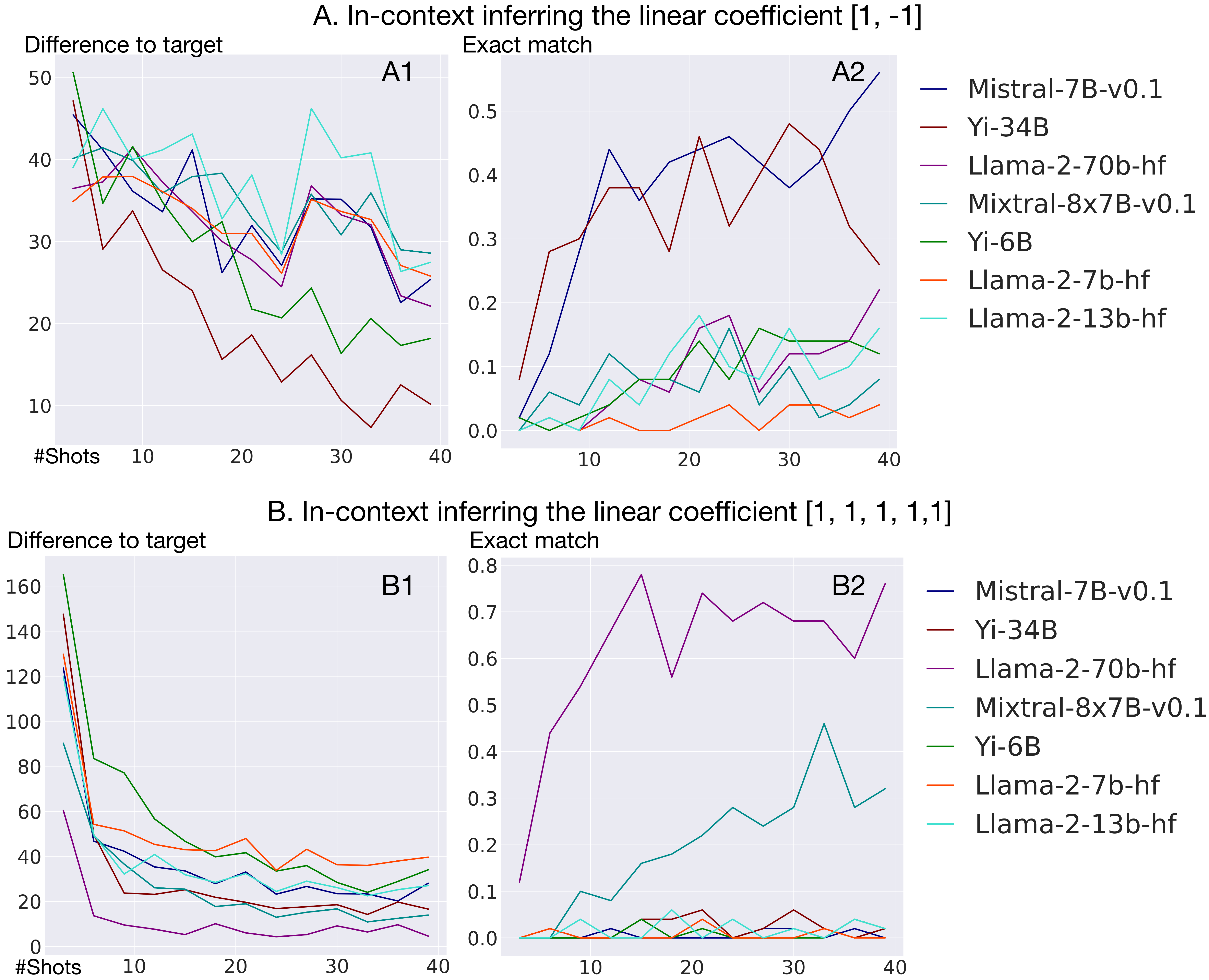}
	\end{center}
	\caption{Evaluating language model's in-context learning capability by inferring the linear coefficients of a weighted sum.
        Considering the discussions of whether emergent ability is an artifact of measurement~\citep{schaeffer2024emergent}, we use 
        difference to the target (target number - model prediction) as a continuous measure, 
        and exact match (target number == model prediction) as a discontinuous measure.
        A: when there is two linear coefficients, Yi-34B performs the best when measuring by the difference to the target number. 
        B: increasing the number of linear coefficients to 5, only models that are large enough (LLaMA2 70B and Mixtral 8x7B) can achieve meaningful exact match, showing that in-context learning complex functions is an emergent ability.
        } %
	\label{fig:icl}
\end{figure}

%% file: 051_eval_chat.tex
In this section, we report the automatic and human preference evaluation of the Chat Model. 
We use greedy decoding to generate responses. For the automatic evaluation benchmarks, we extract answers from the model's generated outputs and calculate accuracy. During the evaluation process, we observed that different prompts have varying influence on  results. Therefore, for the same set of questions, we use identical prompts to evaluate all models, aiming to ensure as fair and unbiased results as possible.

\subsubsection{Automatic Evaluations}


For automatic evaluation, we use the same benchmarks as is for the base model, detailed in Sec. \ref{sec:main results}.
We use both zero-shot and few-shot methods but generally, zero-shot is more suitable for chat models. Our evaluation involves generating responses while following instructions explicitly or implicitly (such as the format in the few-shot examples). We then isolate relevant answers from the generated text. Unlike the base model, for the zero-shot evaluations on the GSM8K and BBH datasets, we employ the Chain-of-Thought (CoT) approach to guide the model in deliberation before reaching an answer.

The results shown in Tab. \ref{tab:chat_models} demonstrate the effectiveness of our chat models in understanding human instructions and generating appropriate instruction-following responses. 
We particularly highlight the 4-bit quantization results, as 4-bit quantization substantially reduces the memory requirement while the model performance nearly does not drop. 
This observation serve as the foundation of serving the model on consumer-grade devices. 


In line with Goodhart's principle, when a measurement metric becomes the target of our pursuit, it ceases to serve as a reliable standard of assessment. Consequently, the outcomes of our evaluations on benchmarks are exclusively employed for ensuring that our alignment training does not detrimentally impact the foundational knowledge and capabilities of the base model. We do not engage in targeted optimization of our chat model with the objective of enhancing benchmark performance.

To further evaluate the generalizability of our model's capabilities, we conducted assessments of its mathematical computation proficiency by subjecting it to the 2023 Hungarian high school mathematics final exam questions, first proposed by the xAI Grok team then reproduced by~\citet{testing_language_models_on_a_held_out_high_school_national_finals_exam}. This evaluation was undertaken with the aim of determining whether our model exhibited signs of overfitting to training datasets that are mathematically oriented.
The results in Fig.~\ref{fig:hungarian_national_hs_finals_exam} show that Yi-34B-Chat performs inspiringly on both the GSM8K and the Hungarian mathematics exam. However, note that Yi-6B-Chat does not exhibit strong mathematical capabilities (on both GSM8K and the Hungarian mathematics exam). We speculate that smaller models may require more data to activate their corresponding abilities during the SFT stage.

\input{tables/tab_chat}

\input{images/fig_math}

\subsubsection{Human Evaluations}


In this section we conducted an assessment of the model's conversational abilities, considering aspects to ensure its effectiveness and safety. We have compiled a collection of open-source evaluation datasets from the community, such as alpaca-eval\cite{dubois2023alpacafarm}, Belle-eval~\citep{belle2023exploring}, and MT-bench\cite{zheng2023judging}. Additionally, we have established our own helpful and harmless evaluation dataset by gathering and constructing data of varying difficulty levels, for the purpose of comprehensively assessing the conversational abilities of chat models.

However, whether it is a public evaluation set or a self-built evaluation set, the evaluation results are strongly influenced by the assessment criteria and the design of the prompt. Our internal evaluation results may be unfair to other models, making it difficult to accurately represent the true capability level of our model.
Therefore, here we only present external evaluation results to demonstrate the current conversational abilities of our chat model.
We consider: 
(1). AlapcaEval\footnote{https://tatsu-lab.github.io/alpaca\_eval/}~\citep{alpaca_eval}, which is designed to assess the English conversation capabilities of models by comparing the responses of a specified model to reference replies from Davinci003~\citep{dubois2023alpacafarm} in order to calculate a win-rate;
(2). LMSys\footnote{https://huggingface.co/spaces/lmsys/chatbot-arena-leaderboard}~\citep{zheng2023judging} Chatbot Arena, which showcases the responses of different models through a dialogue platform, then asks users to make selections based on their preferences, then computes the Elo score; (3). SuperClue\footnote{https://www.superclueai.com/}, on the other hand, is a leaderboard aimed at comprehensively evaluating the Chinese language capabilities of models.

\begin{table*}[t!]
\centering
\scalebox{0.9}{
\begin{tabular}{l c c c c c c c c c c c}

\toprule
\bf Model & \bf Size & \bf AlpacaEval & \bf LMSys Chatbot Arena & \bf SuperClue \\ 



\midrule

GPT-4-Turbo & - &  \textbf{97.7} & \textbf{1243} & \textbf{89.79} \\


GPT-3.5-Turbo & - &  89.37 & 1117 & 59.39 \\


LLaMA2-Chat & 70B &  92.66 & 1077 & - \\


Yi-Chat & 34B &  94.08 & 1110 & 71.87 \\

\bottomrule

\end{tabular}
}
\caption{Human evaluation comparison with other open-source chat models.}
\label{tab:human_evaluation}
\end{table*}

Tab. \ref{tab:human_evaluation} presents the performance results of Yi-34B-Chat in the three third-party evaluations we consider, with the cutoff date for the results being December 21, 2023. The data demonstrates that, although there is still a gap compared to GPT-4, our model exhibits proficient bilingual (Chinese and English) dialogue capabilities and aligns well with user preferences. Additional comparative results of various models are accessible for review on the official website.

\input{images/fig_sftdata}
We further demonstrate the data quality by comparing the speed of preference increase during data scaling.
As is shown in Fig.~\ref{fig:sftdata}, when compared with UltraChat~\citep{ding2023enhancing} and its cleaned version UltraChat 200K, we see a clear tendency of performance improvements when scaling up the Yi data. 

%% file: tables/tab_chat.tex
\begin{table*}[t!]
\centering
\setlength\tabcolsep{2pt}
\resizebox{\textwidth}{!}{
\begin{tabular}{@{}l c c c c c c c c c c c@{}}
\toprule
\multicolumn{1}{l}{\multirow{2}{*}{\textbf{Model}}}  & \multicolumn{1}{c}{\multirow{2}{*}{\textbf{Size}}} & \multicolumn{1}{c}{\multirow{2}{*}{\textbf{MMLU}}}   & \multicolumn{1}{c}{\multirow{2}{*}{\textbf{CMMLU}}} & \multicolumn{1}{c}{\multirow{2}{*}{\textbf{C-Eval(val)}}} & \multicolumn{1}{c}{\multirow{2}{*}{\textbf{TruthfulQA}}} & \multicolumn{1}{c}{\multirow{2}{*}{\textbf{BBH}}} & \multicolumn{1}{c}{\multirow{2}{*}{\textbf{GSM8K}}} \\ 

\multicolumn{1}{c}{} & \multicolumn{1}{c}{} & \multicolumn{1}{c}{} & \multicolumn{1}{c}{}    & \multicolumn{1}{c}{}  & \multicolumn{1}{c}{} & \multicolumn{1}{c}{} &  \multicolumn{1}{c}{} &  \multicolumn{1}{c}{} \\

\multicolumn{1}{c}{} & \multicolumn{1}{c}{} & \multicolumn{1}{c}{\textbf{0-shot / 5-shot}}  & \multicolumn{1}{c}{\textbf{0-shot / 5-shot}} & \multicolumn{1}{c}{\textbf{0-shot / 5-shot}} &  \multicolumn{1}{c}{\textbf{0-shot}} &  \multicolumn{1}{c}{\textbf{0-shot / 3-shot}} & \multicolumn{1}{c}{\textbf{0-shot / 4-shot}}  \\

\midrule

LLaMA2-Chat
& 13B & 50.9 / 47.3 & 27.5 / 35.1 & 27.9 / 35.9 & 36.8 & 32.9 / 58.2 & 36.9 / 2.7 \\
& 70B & 59.4 / 59.9 & 36.1 / 41.0 & 35.0 / 41.3 & 54.0  & 42.4 / 58.5 & 47.1 / 58.7 \\
\gmidrule
Baichuan2-Chat
& 13B & 55.1 / 50.1 & 58.6 / 59.5 & 56.0 / 54.8 & 49.0 & 38.8 / 47.2 & 45.7 / 23.3 \\
\gmidrule
Qwen-Chat
& 14B & 64.0 / 65.0 & 67.7 / 70.6 & 66.1 / 70.1 & 52.5 & 49.7 / 55.0 & 59.5 / 61.2 \\
\gmidrule
InternLM-Chat
& 20B & 55.6 / 57.4 & 53.6 / 53.8 & 51.2 / 53.6 & 51.8 & 42.4 / 36.7 & 15.7 / 43.4 \\
\gmidrule
AquilaChat2
& 34B & 65.2 / 66.7 & 67.5 / 70.0 & \textbf{83.0} / \textbf{89.4} & \textbf{64.3} & 20.1 / 34.3 & 11.5 / 48.5 \\
\gmidrule
Yi-Chat
& 6B & 58.2 / 61.0 & 69.4 / 74.7 & 68.8 / 74.2 & 50.6 & 39.7 / 47.2 & 38.4 / 44.9 \\
Yi-Chat-8bits(GPTQ)
& 6B & 58.3 / 61.0 & 69.2 / 74.7 & 69.2 / 73.9 & 49.9 & 40.4 / 47.3 & 39.4 / 44.9 \\
Yi-Chat-4bits(AWQ)
& 6B & 56.8 / 59.9 & 67.7 / 73.3 & 67.5 / 72.3 & 50.3 & 37.7 / 43.6 & 35.7 / 38.4 \\
\gmidrule
Yi-Chat
& 34B & \textbf{67.6} / 73.5 & \textbf{79.1} / \textbf{81.3} & 77.0 / 78.5 & 62.4 & 51.4 / \textbf{71.7} & \textbf{71.7} / \textbf{76.0} \\
Yi-Chat-8bits(GPTQ)
& 34B & 66.2 / \textbf{73.7} & 79.1 / 81.2 & 76.8 / 79.0 & 61.8 & \textbf{52.1} / 71.0 & 70.7 / 75.7 \\
Yi-Chat-4bits(AWQ)
& 34B & 65.8 / 72.4 & 78.2 / 80.5 & 75.7 / 77.3 & 61.8 & 48.3 / 69.4 & 70.5 / 74.0 \\

\bottomrule

\end{tabular}
}
\caption{Overall performance on automatic benchmarks compared to open-source chat models. We highlight the 4-bit quantization results, as 4-bit quantization substantially reduces the memory requirement while the model performance nearly does not drop. 
This observation serve as the foundation of serving the model on consumer-grade devices, e.g., RTX4090.
}
\label{tab:chat_models}
\end{table*}

%% file: images/fig_math.tex
\begin{figure}
  \centering
  \includegraphics[scale=0.65]{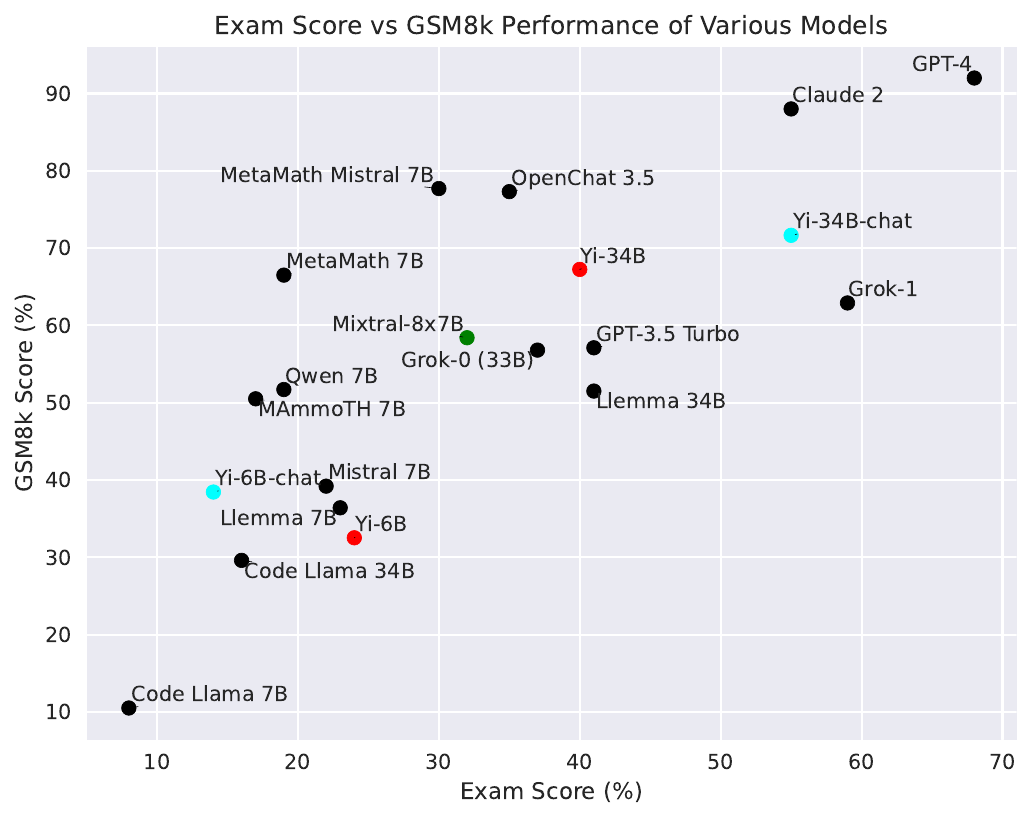}
  \caption{Yi's result of Hungarian mathematics exam. 
  } 
  \label{fig:hungarian_national_hs_finals_exam}
\end{figure}

%% file: images/fig_sftdata.tex
\begin{figure}
	\begin{center}
		\includegraphics[width=\textwidth]{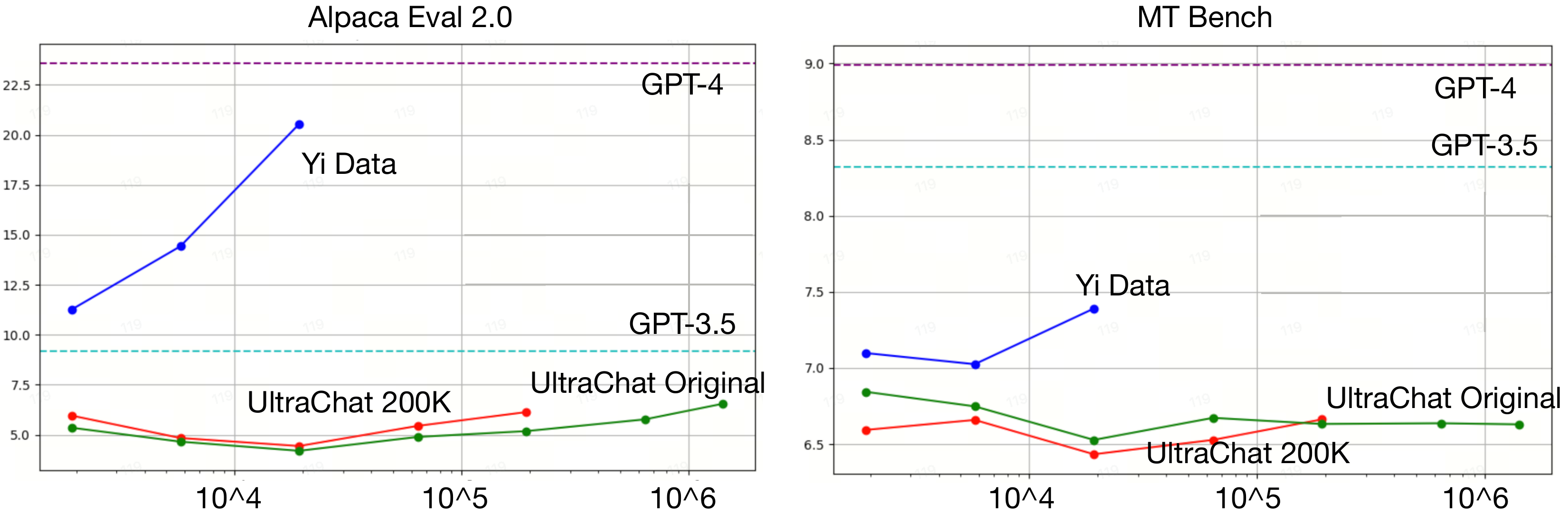}
	\end{center}
	\caption{SFT data scaling curve. Compared with UltraChat and its cleaned version UltraChat 200K, our SFT data demonstrates clear scaling advantages. We attribute its steep slope to the data quality. 
        } %
	\label{fig:sftdata}
\end{figure}

%% file: 021_long_context.tex
Our long-context solution consists of a continual pretraining and a finetuning phase, both are light-weight.
We hold the basic hypothesis that the potential of utilizing information anywhere within the 200K input context is already 
exist in the base model (same as~\citealt{fu2024data}), the continue pretraining phase ``unlocks'' such capability, evidenced by a strong performance on Needle-in-a-Haystack test, then the finetuning phase further adapt the style of response to follow human instruction and preference. 

\textbf{Continue Pretraining}\quad
We continue pretrain the full-attention model using sequence parallelism~\citep{li2021sequence} and distributed attention.
This is to say, we do not use any sparse or linear attention, but use a brute force implementation of the full attention.
We continue pretrain the Yi 6B/ 34B base model on the data mixture of 
(1). original pretraining data, as is introduced in section~\ref{sec:pretrain}; 
(2). length-upsampled long-context data, where the long documents are mostly from books; 
(3). multi-document question-answering synthetic data, where we construct QA pairs where the answer contains a recitation of the related paragraph before the answer. 
Our data approach mostly follows the data engineering practice in~\citet{fu2024data} and~\citet{yu2023paraphrasing}. 
We continue pretrain the model on 5B tokens with 4M batch size, which translate to 100 optimization steps.
Aligning with the concurrent work from~\citet{fu2024data}, we observe that such light-weight continue pretraining is already able to enable a strong performance on Needle-in-a-Haystack test, as we will show in Figure~\ref{fig:needle}. 

\textbf{Supervised Finetuning}\quad
We mix our short-context SFT data with long-context document question-answering data. 
We use model-assisted automated methods (i.e., synthetic data) to construct document QA.
Specifically, we randomly concatenate multiple documents into a sequence, sample one or more paragraphs from the long sequence, and ask a chat model to construct question and answer pairs based on the sampled paragraph.
One important detail is recitation and rephrasing: before giving the answer, we ask the model to recite or paraphrase the original paragraph. 
This data format encourages the model's retrieval behavior and consequently discourages the hallucination behavior: given a question, the model is more likely to use the information within the input to construct the answer, rather than use its internal knowledge, which may be related but inaccurate. 
Our finetuned model is deployed at \url{www.wanzhi01.com}, and we encourage the readers to try it out. 

\input{images/fig_needle}
\input{tables/tab_long_context}
The performance of the 200K models is shown in figure.~\ref{fig:needle} and table~\ref{tab:long_context_mmlu}. 
Specifically, Figure~\ref{fig:needle} shows the famous Needle-in-a-Haystack test of Yi-34B-200K, though we tend to view that this level of retrieval is relatively easy for long-context LLMs. 
Table~\ref{tab:long_context_mmlu} shows that our context scaling does not significantly influence the short-context generic capability.

%% file: images/fig_needle.tex
\begin{figure}
	\begin{center}
		\includegraphics[width=\textwidth]{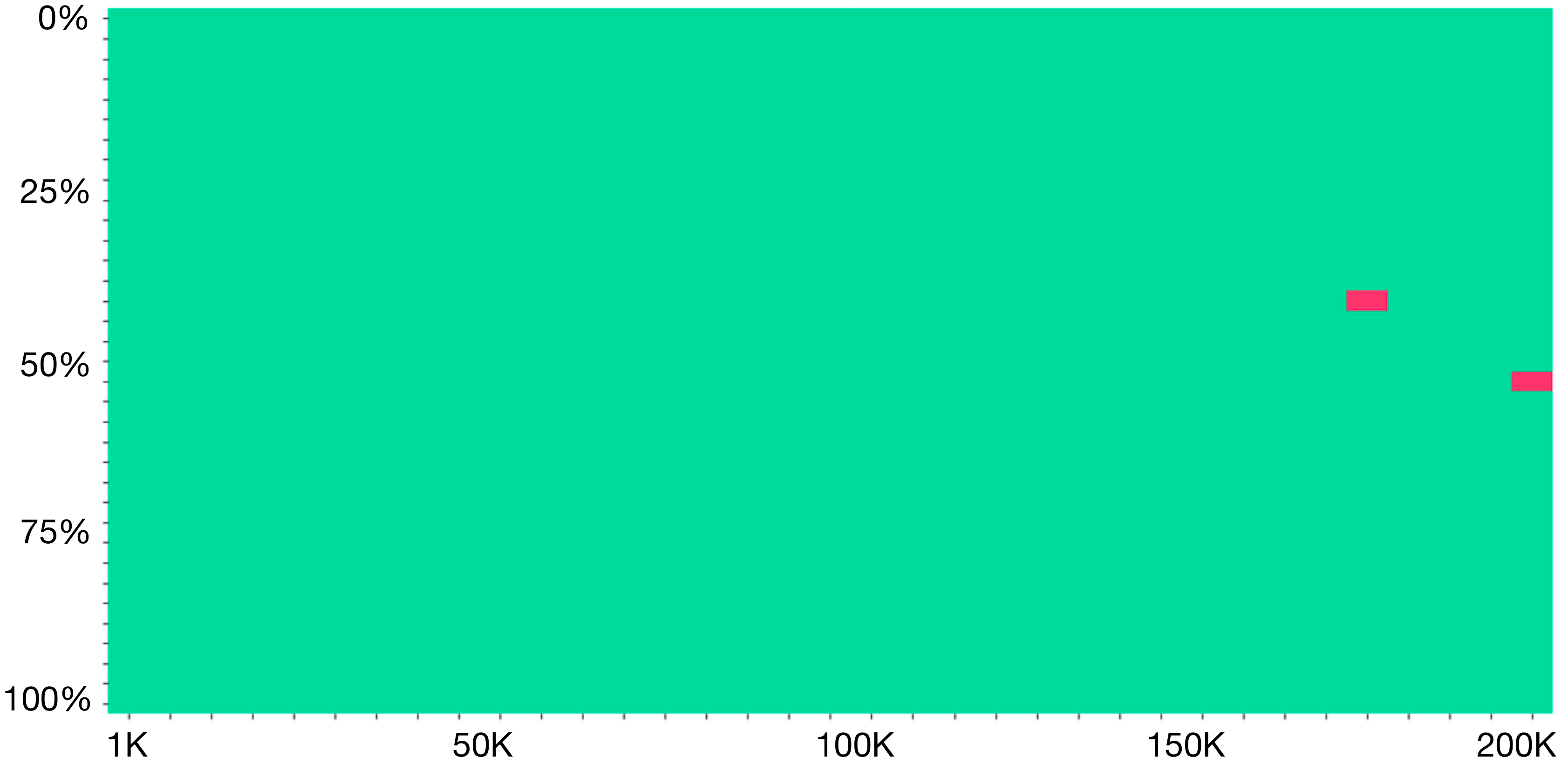}
	\end{center}
	\caption{Needle-in-a-Haystack performance of Yi-34B-200K. 
        X-axis means length of the document, and Y-axis means the depth of the needle sentence within the document. 
        We continue pretrain the model on 5B tokens long-context data mixture and demonstrates a near-all-green performance.
        } %
	\label{fig:needle}
\end{figure}

%% file: tables/tab_long_context.tex
\begin{table}[t!]
  \centering
  \begin{tabular}{@{}lccccc@{}}
\toprule
Model   & \bf Average  & \bf Humanity & \bf STEM  & \bf Social Science & \bf Other \\
\midrule
Yi-6B 4K       & 63.24 & 59.10   & 53.15 & 73.83          & 69.26 \\
Yi-6B 200K     & 61.73 & 56.17   & 52.36 & 72.54          & 68.94 \\
Yi-34B 4K      & 76.32 & 73.17   & 68.03 & 85.11          & 80.78 \\
Yi-34B 200K    & 75.56 & 72.20   & 66.83 & 84.76          & 80.40 \\
\bottomrule

  \end{tabular}
   \caption{Performance on MMLU after 200K adaptation. Extending the context length to 200K does not significantly change the short context capability. 
   }
  \label{tab:long_context_mmlu}
\end{table}

%% file: 022_vision_language.tex
\begin{figure}[t]
    \centering
    \includegraphics[width=0.98\textwidth]{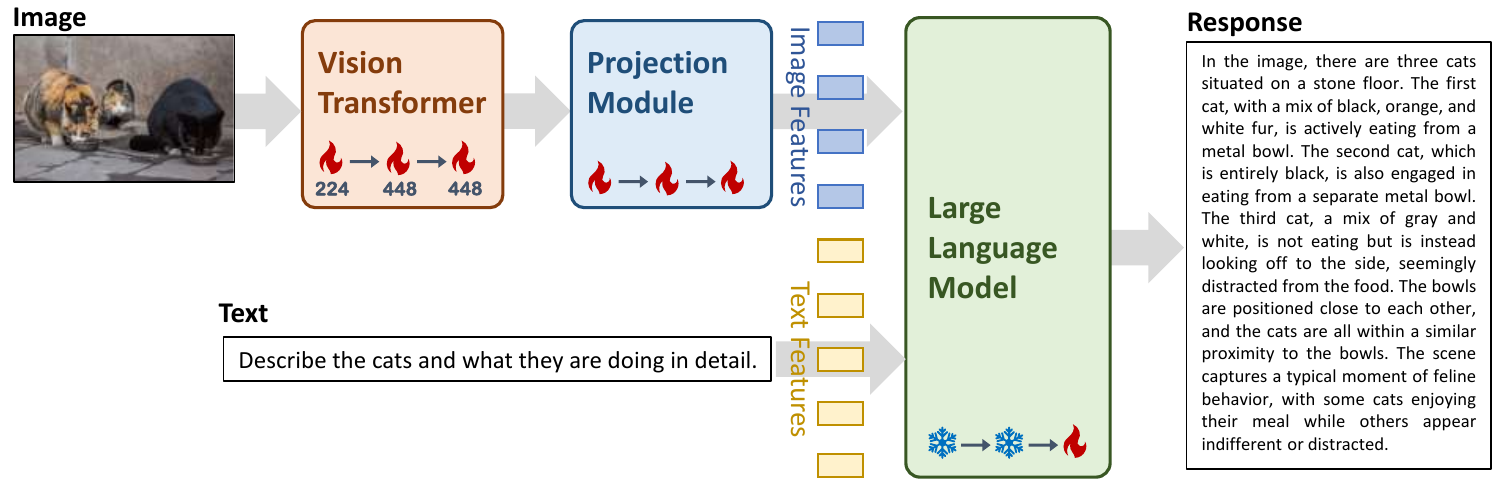}
    \caption{Architecture of Yi-VL models. Symbols are used to denote the training status of various modules at three training stages: a fire icon (\includegraphics[width=1em]{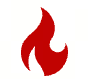}) indicates the parameters of the module are trainable, while a snowflake icon (\includegraphics[width=0.9em]{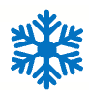}) signifies that parameters are frozen. The image resolution used in ViT at each stage, either $224^2$ or $448^2$, is also marked.}
    \label{fig:yi-vl}
\end{figure}

In the burgeoning field of multimodal research, the integration of image understanding capabilities into large language models has become increasingly viable. 
Drawing inspiration from the open-sourced LLaVA~\citep{liu2023visual,liu2023improved}, we present Yi Vision Language (Yi-VL) models, \textit{i.e.}, Yi-VL-6B and Yi-VL-34B, based on Yi-6B-Chat and Yi-34B-Chat language models. 
The architecture of Yi-VL models, as illustrated in Figure \ref{fig:yi-vl}, comprises three primary modules. The Vision Transformer (ViT), used for image encoding, is initialized with CLIP ViT-H/14 model~\citep{ilharco_gabriel_2021_5143773}. A Projection Module, designed to align image features with text feature spcae, consists of a two-layer Multilayer Perceptron (MLP) with layer normalizations. Finally, the large language model, initialized with the Yi-Chat models, demonstrating exceptional proficiency in understanding and generating both English and Chinese. To enhance the performance of Yi-VL models in bilingual multimodal understanding and generation, we leverage a rich dataset of bilingual image-text pairs.

Yi-VL models undergo a three-stage training process: 
\begin{description}[leftmargin=0.2in]
    \item[Stage 1:] we train the parameters of the ViT and the projection module using an image resolution of $224^2$. The training leverages a substantial dataset comprising $100$ million image-text pairs from LAION-400M~\citep{schuhmann2021laion}. The primary objective is to enhance the ViT's knowledge acquisition within our specified architecture and to achieve better alignment between the ViT and the LLM. 
    \item[Stage 2:] we scale up the image resolution of ViT to $448^2$, aiming to further boost the model's capability for discerning intricate visual details. The dataset used in this stage includes $20$ million image-text pairs derived from LAION-400M. Additionally, we incorporate around $4.8$ million image-text pairsn from diverse sources, \textit{e.g.}, CLLaVA~\citep{linksoulai2023chinesellava}, LLaVAR~\citep{zhang2023llavar}, Flickr~\citep{young2014image}, VQAv2~\citep{goyal2017making}, RefCOCO~\citep{kazemzadeh2014referitgame}, Visual7w~\citep{zhu2016visual7w} and so on.
    \item[Stage 3:] the parameters of the entire model are trained. The primary goal is to enhance the model's proficiency in multimodal chat interactions, thereby endowing it with the ability to seamlessly integrate and interpret visual and linguistic inputs. To this end, the training dataset encompasses a diverse range of sources, totalling approximately $1$ million image-text pairs, including GQA~\citep{hudson2019gqa}, VizWiz VQA~\citep{gurari2018vizwiz}, TextCaps~\citep{sidorov2020textcaps}, OCR-VQA~\citep{mishra2019ocr}, Visual Genome~\citep{krishna2017visual}, ShareGPT4V~\citep{chen2023sharegpt4v} and so on. To ensure data balancing, we impose a cap on the maximum data contribution from any single source, restricting it to no more than $50,000$ pairs. 
\end{description}

In Stage 1 and 2, we set the global batch size, the learning rate, the gradient clip and the number of epoch to $4096$, $1$e$-4$, $0.5$ and $1$, respectively. In Stage 3, these parameters are adjusted to $256$, $2$e$-5$, $1.0$ and $2$. The training consumes $128$ NVIDIA A100 GPUs. The total training time amounted to approximately $3$ days for Yi-VL-6B and $10$ days for Yi-VL-34B. 

\input{tables/tab_yi_vl}

Table~\ref{tab:yi_vl_mmmu} shows the MMMU test set leaderboard by Yi-VL's release. We note that this area is currently actively under research, aligning with the community's advances, we will continuously improve the update Yi-VL's performance. 

%% file: tables/tab_yi_vl.tex
\begin{table}[t!]
\centering
\begin{tabular}{@{}lccccccc@{}}
\toprule
\textbf{Model}          & \textbf{Overall} & \textbf{Art} & \textbf{Business} & \textbf{Science} & \textbf{Health} & \textbf{Society} & \textbf{Engineering} \\ \midrule
GPT-4V & 55.7             & 65.3                   & 64.3              & 48.4             & 63.5                        & 76.3                         & 41.7                   \\
Yi-VL-34B               & 41.6             & 56.1                   & 33.3              & 32.9             & 45.9                        & 66.5                         & 36.0                   \\
Qwen-VL-PLUS            & 40.8             & 59.9                   & 34.5              & 32.8             & 43.7                        & 65.5                         & 32.9                   \\
Marco-VL                & 40.4             & 56.5                   & 31.0              & 31.0             & 46.9                        & 66.5                         & 33.8                   \\
Yi-VL-6B                & 37.8             & 53.4                   & 30.3              & 30.0             & 39.3                        & 58.5                         & 34.1                   \\
InfMIM-Zephyr-7B        & 35.5             & 50.0                   & 29.6              & 28.2             & 37.5                        & 54.6                         & 31.1                   \\
SVIT                    & 34.1             & 48.9                   & 28.0              & 26.8             & 35.5                        & 50.9                         & 30.7                   \\
Emu2-Chat               & 34.1             & 50.6                   & 27.7              & 28.0             & 32.4                        & 50.3                         & 31.3                   \\
BLIP-2 FLAN-T5-XXL       & 34.0             & 49.2                   & 28.6              & 27.3             & 33.7                        & 51.5                         & 30.4                   \\
InstructBLIP-T5-XXL      & 33.8             & 48.5                   & 30.6              & 27.6             & 33.6                        & 49.8                         & 29.4                   \\
LLaVA-1.5-13B            & 33.6             & 49.8                   & 28.2              & 25.9             & 34.9                        & 54.7                         & 28.3                   \\
Qwen-VL-7B-Chat          & 32.9             & 47.7                   & 29.8              & 25.6             & 33.6                        & 45.3                         & 30.2                   \\
SPHINX*                  & 32.9             & 50.9                   & 27.2              & 25.3             & 34.1                        & 51.2                         & 27.8                   \\
mPLUG-OWL2              & 32.1             & 48.5                   & 25.6              & 24.9             & 32.8                        & 46.7                         & 29.6                   \\
BLIP-2 FLAN-T5-XL        & 31.0             & 43.0                   & 25.6              & 25.1             & 31.8                        & 48.0                         & 27.8                   \\
InstructBLIP-T5-XL       & 30.6             & 43.3                   & 25.2              & 25.2             & 29.3                        & 45.8                         & 28.6                   \\
CogVLM                   & 30.1             & 38.0                   & 25.6              & 25.1             & 31.2                        & 41.5                         & 28.9                   \\ \bottomrule
\end{tabular}
\caption{\label{tab:yi_vl_mmmu}MMMU test set
 performance by the time of Yi-VL's release.}
\end{table}

%% file: 023_depth_upscaling.tex
Recent studies on scaling laws~\citep{henighan2020scaling, hoffmann2022training, kaplan2020scaling} have underscored the predictable improvement in model performance with increases in computational budget, model size, and data size. Yet, identifying the most effective distribution of resources between model and data sizes upon expanding the computational budget remains a formidable challenge in the field of scaling laws. Additionally, research conducted by~\citet{deepseekai2024deepseek} has highlighted that the allocation of an increased computational budget towards model scaling should be proportional to the quality of the data available. In light of these insights, we propose a novel approach aimed at dynamically adjusting the resource allocation between data and model sizes through a series of staged training processes. This strategy iteratively fine-tunes the balance between data characteristics and model size according to scaling laws, enhancing both model training efficiency and performance.

\textbf{Method}\quad
Following the methodology outlined by~\citet{kim2023solar}, our goal is to upscale our Yi-6B base model, which has 32 layers, to a 9B model named the Yi-9B base model, featuring 48 layers, by duplicating the original 16 middle layers 12-28. Depth up-scaling involves expanding the base model's depth and subsequently continuing the pretraining phase for the enhanced model.

\input{images/fig_layer_cosine_graph}
Our investigations reveal that the decision on which layers to replicate could be informed by evaluating the cosine similarity scores between the inputs and outputs of each layer. Such an approach allows for targeted model scaling without necessitating additional pretraining, leading only to minimal performance impacts. This minimal impact on performance is attributed to the high cosine similarity, approaching one, between the inputs and outputs of the duplicated layers, as evidenced in Figure \ref{fig:layer_cosine_score}. This observation suggests that the replication of these layers does not significantly alter the output logits produced by the original model. This method ensures the efficient scaling of the model by optimizing its architecture based on the internal processing dynamics of its layers.

\input{tables/table_depth_upscaling}
\textbf{Continual Training}\quad
The dataset is composed of approximately 800 billion tokens across two stages, with around 70\% having been recently collected and carefully selected. We have enhanced the code coverage in the final stage to improve code performance.

To optimize the training process, we maintain a constant learning rate of 3e-5, and adopt a strategic approach to gradually increase the batch size from 4M tokens whenever the model's loss plateaued. This incremental adjustment of the batch size, alongside maintaining all other parameters in alignment with the established Yi-6B base model configuration, was instrumental in navigating the challenges of training at scale.

The effectiveness of these strategies is demonstrated in Table \ref{tab:depth_upscaling}, which details the Yi-9B base model's performance across a variety of benchmarks, including common sense, reasoning, knowledge, coding, and mathematics. It underscores the competitive advantages of Yi-9B base model in specific domains, illustrating the efficacy of our methodology in enhancing model performance by optimally adjusting the interplay between data characteristics and model size.

%% file: images/fig_layer_cosine_graph.tex
\begin{figure}[t]
  \centering
  \includegraphics[width=0.9\textwidth]{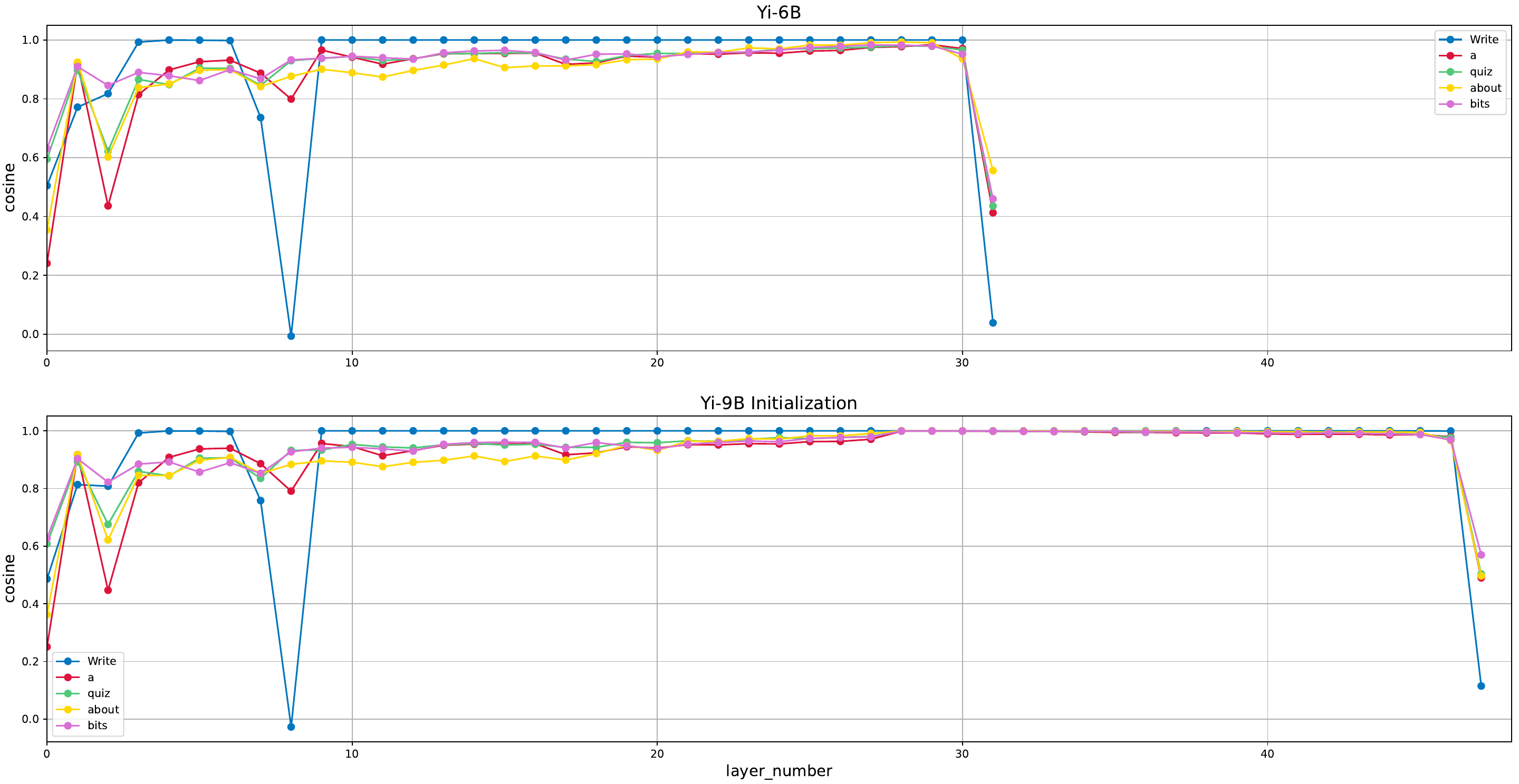}
  \caption{Input/output cosine similarity score of each token per layer for text "Write a quiz about bits". The cosine similarity scores of the 16 newly added layers(layers 28-44), as depicted in the lower figure, are observed to be nearly 1.} 
  \label{fig:layer_cosine_score}
\end{figure}

%% file: tables/table_depth_upscaling.tex
\begin{table*}[tb!]
\centering
\scalebox{0.8}{
\begin{tabular}{lcccccccc}
\toprule
\textbf{Model} & \textbf{Arc-C} & \textbf{HellaSwag} & \textbf{MMLU} & \textbf{Winogrande} & \textbf{GSM8K} & \textbf{MATH} & \textbf{HumanEval} & \textbf{MBPP} \\
\midrule
\textbf{Yi-6B} & 50.3 & 74.4 & 63.2 & 71.3 & 32.5 & 4.6 & 15.9 & 26.3 \\
\textbf{Yi-9B Init} & 52.1 & 73.3 & 63.0 & 69.4 & 31.3 & 4.1 & 12.8 & 25.8  \\
\textbf{Yi-9B} & \textbf{55.6}  & \textbf{76.4} & \textbf{68.4} & \textbf{73.0} & \textbf{52.3} & \textbf{15.9} & \textbf{39.0} & \textbf{54.4} \\
\bottomrule
\end{tabular}
}
\caption{Performance between Yi-6B and Yi-9B: Arc Challenge (25-shot), HellaSwag (10-shot) MMLU (5-shot), Winogrande (5-shot), GSM8K (5-shot), MATH (4-shot), HumanEval pass@1, MBPP pass@1(3-shot). Yi-9B Init is just depthwise upscaling from Yi-6B by duplicating layers 12-28 without further training.}
\label{tab:depth_upscaling}
\end{table*}

%% file: 070_conclusion.tex
In this report, we discuss the full-stack development of the Yi language model family. 
Yi-34B achieves GPT-3.5 matching performance and is deployable (thank to the 4/8-bit quantization) on consumer-grade devices, making it an ideal model for local deployment. 

The key takeaways from the Yi pretraining procedure are about data quantity and quality: 
(1). training the model on a larger amount of data than the Chinchilla optimal delivers clear and consistent performance gain, which we highly recommend for all pretraining teams. 
Our model is trained on 3.1T tokens, yet we belive with larger amount of data, we can continue improve the model performance (i.e., the model have not saturated at 3.1T);
(2). when it comes to the pretraining data quality, we believe the most critical two factors are the source of the data (e.g., whether the text is produced for professional usage or for casual social media posting) and the details of the data cleaning (e.g., the strength of filtering and deduplication). Since data cleaning is a very complicated pipeline and it is extremely difficult to conduct extensive grid-search styled optimizations, our current solution may still have room for improvements. 

The key takeaways from the Yi finetuning procedure is to heavily iterate on a small amount of data ($\le$ 10K), case by case, over multiple iterations, directly by the machine learning engineer, and improved from real user feedback. 
This approach clearly deviates from the instruction-scaling approach, initially introduced by the FLAN series~\citep{chung2022scaling} then followed by the UltraChat series~\citep{ding2023enhancing}. 



As is demonstrated by our current results, the reasoning capability, which we view as the core capability for real-world deployment of language models, is strongly correlated with model scale when the amount of pretraining data is fixed. 
We believe that given our current
results, continuing to scale up model parameters using thoroughly optimized data
will lead to even stronger frontier models in our upcoming next versions.

%% file: app_author_list.tex
Our team members contribute to the development of Yi from the following perspectives: 

\begin{multicols}{2}
\begin{itemize}
    \item Frontier Research
    \item Machine Learning Infrastructure
    \item Pretraining
    \item Finetuning and AI Alignment
    \item Multimodal
    \item Safety and Responsible AI 
    \item Deployment
\end{itemize}
\end{multicols}

We list our team members in alphabetical order. All authors contributed equally to this work. 

\begin{multicols}{2}
\begin{itemize}
\item Alex Young
\item Bei Chen
\item Chao Li
\item Chengen Huang
\item Ge Zhang
\item Guanwei Zhang
\item Guoyin Wang
\item Heng Li
\item Jiangcheng Zhu
\item Jianqun Chen
\item Jing Chang
\item Kaidong Yu
\item Peng Liu
\item Pengcheng Nie
\item Qiang Liu
\item Shawn Yue
\item Senbin Yang
\item Shiming Yang
\item Wen Xie
\item Wenhao Huang
\item Xiaohui Hu
\item Xiaoyi Ren
\item Xinyao Niu
\item Yanpeng Li
\item Yuchi Xu
\item Yudong Liu
\item Yue Wang
\item Yuxuan Cai
\item Zhenyu Gu
\item Zhiyuan Liu
\item Zonghong Dai
\end{itemize}
\end{multicols}